%% file: eacl2021.tex
\pdfoutput=1
\documentclass[11pt,a4paper]{article}
\usepackage[hyperref]{eacl2021}
\usepackage{times}
\usepackage{latexsym}

\usepackage{microtype}

\usepackage{enumitem}
\usepackage{booktabs}
\usepackage{tikz}
\usetikzlibrary{arrows.meta, matrix, automata, fit, shapes, positioning, calc, intersections}

\newcommand{\linestack}[1]{\def\arraystretch{0.8}\begin{tabular}[c]{@{}c@{}} #1 \end{tabular}}

\aclfinalcopy 
\def\aclpaperid{661} 

\title{Benchmarking Machine Reading Comprehension: \\ A Psychological Perspective}

\author{Saku Sugawara$^1$, Pontus Stenetorp$^2$, Akiko Aizawa$^1$ \\
  $^1$National Institute of Informatics, $^2$University College London \\
  {\tt \{saku,aizawa\}@nii.ac.jp},~{\tt~p.stenetorp@cs.ucl.ac.uk}\\
}

\date{}

\newcommand{\todo}[1]{\colorbox{red!30}{TODO}}
\newcommand{\toread}[1]{}

\begin{document}
\maketitle

\begin{abstract}
  Machine reading comprehension (MRC) has received considerable attention as a benchmark for natural language understanding.
  However, the conventional task design of MRC lacks explainability beyond the model interpretation, i.e., reading comprehension by a model cannot be explained in human terms.
  To this end, this position paper provides a theoretical basis for the design of MRC datasets based on psychology as well as psychometrics, and summarizes it in terms of the prerequisites for benchmarking MRC.
  We conclude that future datasets should (i) evaluate the capability of the model for constructing a coherent and grounded representation to understand context-dependent situations and (ii) ensure substantive validity by shortcut-proof questions and explanation as a part of the task design.
\end{abstract}

\section{Introduction}
\label{sec:intro}

\begin{table*}[t]
  \centering
  {\small
  \begin{tabular}{p{8em}p{10em}p{16em}p{10em}} \toprule
      Question & Foundation & Requirements & Future direction \\ \midrule
      What is reading comprehension? &
      Representation levels in human reading comprehension: (A) surface structure, (B) textbase, and (C) situation model. &
      (A) Linguistic-level sentence understanding, (B) comprehensiveness of skills for inter-sentence understanding, and (C) evaluation of coherent representation grounded to non-textual information. &
      (C) Dependence of context on defeasibility and novelty, and grounding to non-textual information with a long passage. \\ \midrule
      How can we evaluate reading comprehension? &
      Construct validity in psychometrics: (1) content, (2) substantive, (3) structural, (4) generalizability, (5) external, and (6) consequential aspects. &
      (1) Wide coverage of skills, (2) evaluation of the internal process, (3) structured metrics, (4) reliability of metrics, (5) comparison with external variables, and (6) robustness to adversarial attacks and social biases. &
      (2)~Creating shortcut-proof questions by filtering and ablation, and designing a task for validating the internal process. \\
      \bottomrule
  \end{tabular}}
  \caption{Overview of theoretical foundations, requirements, and future directions of MRC discussed in this paper.}
  \label{tab:overview}
\end{table*}

Evaluation of natural language understanding (NLU) is a long-standing goal in the field of artificial intelligence.
Machine reading comprehension (MRC) is a task that tests the ability of a machine to read and understand unstructured text and could be the most suitable task for evaluating NLU because of its generic formulation \cite{chen2018neural}.
Recently, many large-scale datasets have been proposed, and deep learning systems have achieved human-level performance for some of these datasets.

However, analytical studies have shown that MRC models do not necessarily achieve human-level understanding.
For example, \citet{jia2017adversarial} use manually crafted adversarial examples to show that successful systems are easily distracted.
\citet{sugawara2020assessing} show that a significant part of already solved questions is solvable even after shuffling the words in a sentence or dropping content words.
These studies demonstrate that we cannot explain what type of understanding is required by the datasets and is actually acquired by models.
Although benchmarking MRC is related to the intent behind questions and is critical to test hypotheses 
from a top-down viewpoint \cite{bender2020climbing}, its theoretical foundation is poorly investigated in the literature.

In this position paper, we examine the prerequisites for benchmarking MRC based on the following two questions:
(i) \emph{What} does reading comprehension involve?
(ii) \emph{How} can we evaluate it?
Our motivation is to provide a theoretical basis for the creation of MRC datasets. 
As \citet{gilpin2018explaining} indicate, interpreting the internals of a system is closely related to only the system's architecture and is insufficient for explaining how the task is accomplished.
This is because even if the internals of models can be interpreted, we cannot explain what is \emph{measured} by the datasets.
Therefore, our study focuses on the explainability of the task rather than the interpretability of models.

We first overview MRC and review the analytical literature that indicates that existing datasets might fail to correctly evaluate their intended behavior (Section~\ref{sec:overview}).
Subsequently, we present a psychological study of human reading comprehension in Section~\ref{sec:what-mrc} for answering the \emph{what} question.
We argue that the concept of \emph{representation levels} can serve as a conceptual hierarchy for organizing the technologies in MRC.
Section~\ref{sec:how-mrc} focuses on answering the \emph{how} question.
Here, we implement psychometrics to analyze the prerequisites for the task design of MRC.
Furthermore, we introduce the concept of \emph{construct validity}, which emphasizes validating the interpretation of the task's outcome.
Finally, in Section~\ref{sec:direction}, we explain the application of the proposed concepts into practical approaches, highlighting potential future directions toward the advancement of MRC.
Regarding the \emph{what} question, we indicate that datasets should evaluate the capability of the \emph{situation model}, which refers to the construction of a coherent and grounded representation of text based on human understanding.
Regarding the \emph{how} question, we argue that among the important aspects of the construct validity, \emph{substantive validity} must be ensured, which requires the verification of the internal mechanism of comprehension.

\toread\ Table~\ref{tab:overview} provides an overview of the perspectives taken in this paper.
Our answers and suggestions to the \emph{what} and \emph{how} questions are summarized as follows:
(1) Reading comprehension is the process of creating a situation model that best explains given texts and the reader's background knowledge.
The situation model should be the next focal point in future datasets for benchmarking the human-level reading comprehension.
(2) To evaluate reading comprehension correctly, the task needs to provide a rubric (scoring guide) for sufficiently covering the aspects of the construct validity.
In particular, the substantive validity should be ensured by creating shortcut-proof questions and by designing a task formulation that is explanatory itself.

\section{Task Overview}
\label{sec:overview}

\subsection{Task Variations and Existing Datasets}

MRC is a task in which a machine is given a document (\emph{context}) and it answers the questions based on the context.
\citet{burges2013towards} provides a general definition of MRC, i.e., \emph{a machine comprehends a passage of text if, for any question regarding that text that can be answered correctly by a majority of native speakers, that machine can provide a string which those speakers would agree both answers that question}.
We overview various aspects of the task along with representative datasets as follows.
Existing datasets are listed in Appendix \ref{app:datasets}.

\paragraph{Context Styles}
A context can be given in various forms with different lengths such as a single passage (MCTest \cite{richardson2013mctest}), a set of passages (HotpotQA \cite{yang2018hotpotqa}), a longer document (CBT \cite{hill2015goldilocks}), or open domain \cite{chen2017reading}.
In some datasets, a context includes non-textual information such as images (RecipeQA \cite{yagcioglu2018recipeqa}).

\paragraph{Question Styles}
A question can be an interrogative sentence (in most datasets), a fill-in-the-blank sentence (cloze) (CLOTH \cite{xie2018large}), knowledge base entries (QAngaroo \cite{welbl2018constructing}) and search engine queries (MSMARCO \cite{nguyen2016msmarco}).

\paragraph{Answering Styles}
An answer can be (i) chosen from a text span of the given document (\emph{answer extraction}) (NewsQA \cite{trischler2017newsqa}), (ii) chosen from a candidate set of answers (\emph{multiple choice}) (MCTest \cite{richardson2013mctest}), or (iii) generated as a free-form text (\emph{description}) (NarrativeQA \cite{kocisky2018narrativeqa}).
Some datasets optionally allow answering by a \emph{yes}/\emph{no} reply (BoolQ \cite{clark2019boolq}).

\paragraph{Sourcing Methods}
Initially, questions in small-scale datasets are created by experts (QA4MRE \cite{sutcliffe2013QA4MRE}).
Later, fueling the development of neural models, most published datasets have more than a hundred thousand questions that are automatically created (CNN/Daily Mail \cite{hermann2015teaching}), crowdsourced (SQuAD v1.1 \cite{rajpurkar2016squad}), and collected from examinations (RACE \cite{lai2017race}).

\paragraph{Domains}
The most popular domain is Wikipedia articles (Natural Questions \cite{kwiatkowski2019natural}), but news articles are also used (Who-did-What \cite{onishi2016who}).
CliCR \cite{suster2018clicr} and emrQA \cite{pampari2018emrqa} are datasets in the clinical domain.
DuoRC \cite{saha2018duorc} uses movie scripts.

\paragraph{Specific Skills}
Several recently proposed datasets require specific skills including unanswerable questions (SQuAD v2.0 \cite{rajpurkar2018know}), dialogues (CoQA \cite{reddy2019coqa}, DREAM \cite{sun2019dream}), multiple-sentence reasoning (MultiRC \cite{khashabi2018looking}), multi-hop reasoning (HotpotQA \cite{yang2018hotpotqa}), mathematical and set reasoning (DROP \cite{dua2019drop}), commonsense reasoning (CosmosQA \cite{huang2019cosmos}), coreference resolution (QuoRef \cite{dasigi2019quoref}), and logical reasoning (ReClor \cite{yu2020reclor}).

\subsection{Benchmarking Issues}
\label{sec:issues}

In some datasets, the performance of machines has already reached human-level performance.
However, \citet{jia2017adversarial} indicate that models can easily be fooled by manual injection of distracting sentences.
Their study revealed that questions simply gathered by crowdsourcing without careful guidelines or constraints are insufficient to evaluate precise language understanding.

This argument is supported by further studies across a variety of datasets.
For example, \citet{min2018efficient} find that more than 90\% of the questions in SQuAD \cite{rajpurkar2016squad} require obtaining an answer from a single sentence despite being provided with a passage.
\citet{sugawara2018what} show that large parts of twelve datasets are easily solved only by looking at a few first question tokens and attending the similarity between the given questions and the context.
Similarly, \citet{feng2018pathologies} and \citet{mudrakarta2018did} demonstrate that models trained on SQuAD do not change their predictions even when the question tokens are partly dropped.
\citet{kaushik2018how} also observe that question- and passage-only models perform well for some popular datasets.
\citet{min2019compositional} and \citet{chen2019understanding} concurrently indicate that for multi-hop reasoning datasets, the questions are solvable only with a single paragraph and thus do not require multi-hop reasoning over multiple paragraphs.
\citet{zellers2019hellaswag} report that their dataset unintentionally contains stylistic biases in the answer options which are embedded by a language-based model.

Overall, these investigations highlight a grave issue of the task design, i.e., even if the models achieve human-level accuracies, we cannot prove that they successfully perform reading comprehension.
This issue may be attributed to the low interpretability of black-box neural network models.
However, a problem is that we cannot explain what is measured by the datasets even if we can interpret the internals of models.
We speculate that this benchmarking issue in MRC can be attributed to the following two points:
(i) we do not have a comprehensive theoretical basis of reading comprehension for specifying what we should ask (Section~\ref{sec:what-mrc}) and
(ii) we do not have a well-established methodology for creating a dataset and for analyzing a model based on it (Section~\ref{sec:how-mrc}).\footnote{
  These two issues loosely correspond to the plausibility and faithfulness of explanation \cite{jacovi2020towards}.
  The plausibility is linked to what we expect as an explanation, whereas the faithfulness refers to how accurately we explain models' reasoning process.
}
In the remainder of this paper, we argue that these issues can be addressed by using insights from the psychological study of reading comprehension and by implementing psychometric means of validation.

\section{Reading Comprehension from Psychology to MRC}
\label{sec:what-mrc}

\subsection{Computational Model in Psychology}

Human text comprehension has been studied in psychology for a long time \cite{kintsch2005comprehension,graesser1994constructing,kintsch1988role}.
Connectionist and computational architectures have been proposed for such comprehension including a mechanism pertinent to knowledge activation and memory storing.
Among the computational models, the construction--integration (CI) model is the most influential and provides a strong foundation of the field \cite{mcnamara2009toward}.

The CI model assumes three different representation levels as follows:

\begin{itemize}[leftmargin=1.5em] 
\item \emph{Surface structure} is the linguistic information of particular words, phrases, and syntax obtained by decoding the raw textual input.
\item \emph{Textbase} is a set of propositions in the text, where the propositions are locally connected by inferences (\emph{microstructure}).
\item \emph{Situation model} is a situational and coherent mental representation in which the propositions are globally connected (\emph{macrostructure}), and it is often grounded to not only texts but also to sounds, images, and background information.
\end{itemize}

The CI model first decodes textual information (i.e., the surface structure) from the raw textual input, then creates the propositions (i.e., textbase) and their local connections occasionally using the reader's knowledge (\emph{construction}), and finally constructs a coherent representation (i.e., situation model) that is organized according to five dimensions including time, space, causation, intentionality, and objects \cite{zwaan1998situation}, which provides a global description of the events (\emph{integration}).
These steps are not exclusive, i.e., propositions are iteratively updated in accordance with the surrounding ones with which they are linked.
Although the definition of successful text comprehension can vary, \citet{hernandez2017measure} indicates that comprehension implies the process of creating (or searching for) a situation model that best explains the given text and the reader's background knowledge \cite{zwaan1998situation}.
\toread\ We use this definition to highlight that the creation of a situation model plays a vital role in human reading comprehension.

Our aim in this section is to provide a basis for explaining what reading comprehension is, which requires terms for explanation.
In the computational model above, the representation levels appear to be useful for organizing such terms.
We ground existing NLP technologies and tasks to different representation levels in the next section.

\subsection{Skill Hierarchy for MRC}
\label{sec:hierarchy}

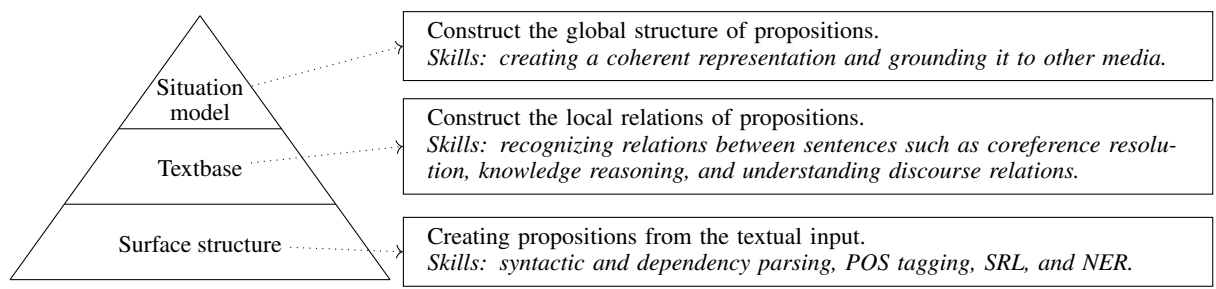
\begin{figure*}[th]
  \centering
  \input{skill-hierarchy}
  \caption{Representation levels and corresponding natural language understanding skills.}
  \label{fig:skill}
\end{figure*}

Here, we associate the existing NLP tasks with the three representation levels introduced above.
The biggest advantage of MRC is its general formulation, which makes it the most general task for evaluating NLU.
This emphasizes the importance of the requirement of various \emph{skills} in MRC, which can serve as the units for the explanation of reading comprehension.
Therefore, our motivation is to provide an overview of the skills as a hierarchical taxonomy and to highlight the missing aspects in existing MRC datasets that are required for comprehensively covering the representation levels.

\paragraph{Existing Taxonomies}
We first provide a brief overview of the existing taxonomies of skills in NLU tasks.
For recognizing textual entailment \cite{dagan2006pascal}, several studies present a classification of reasoning and commonsense knowledge \cite{bentivogli2010building,sammons2010ask,lobue2011commonsense}.
For scientific question answering, \citet{jansen2016whats} categorize knowledge and inference for an elementary-level dataset.
Similarly, \citet{boratko2018systematic} propose types of knowledge and reasoning for scientific questions in MRC \cite{clark2018think}.
A limitation of both these studies is that the proposed sets of knowledge and inference are limited to the domain of elementary-level science.
Although some existing datasets for MRC have their own classifications of skills, they are coarse and only cover a limited extent of typical NLP tasks (e.g., word matching and paraphrasing).
In contrast, for a more generalizable definition, \citet{sugawara2017evaluation} propose a set of 13 skills for MRC.
\citet{rogers2020getting} pursue this direction by proposing a set of questions with eight question types.
In addition, \citet{schlegel2020framework} propose an annotation schema to investigate requisite knowledge and reasoning.
\citet{dunietz2020test} propose a template of understanding that consists of spatial, temporal, causal, and motivational questions to evaluate precise understanding of narratives with reference to human text comprehension.

In what follows, we describe the three representation levels that basically follow the three representations of the CI model but are modified for MRC.
The three levels are shown in Figure~\ref{fig:skill}.
We emphasize that we do not intend to create exhaustive and rigid definitions of skills.
Rather, we aim to place them in a hierarchical organization, which can serve as a foundation to highlight the missing aspects in the current MRC.

\paragraph{Surface Structure}
This level broadly covers the linguistic information and its semantic meaning, which can be based on the raw textual input.
Although these features form a proposition according to psychology, it should be viewed as sentence-level semantic representation in computational linguistics.
This level includes part-of-speech tagging, syntactic parsing, dependency parsing, punctuation recognition, named entity recognition (NER), and semantic role labeling (SRL).
Although these basic tasks can be accomplished by some recent pretraining-based neural language models \cite{liu2019linguistic}, they are hardly required in NLU tasks including MRC.
In the natural language inference task, \citet{mccoy2019right} indicate that existing datasets (e.g., \citet{bowman2015large}) may fail to elucidate the syntactic understanding of given sentences.
Although it is not obvious that these basic tasks should be included in MRC and it is not easy to circumscribe linguistic knowledge from concrete and abstract knowledge \cite{zaenen2005local,manning2006local}, we should always care about the capabilities of basic tasks (e.g., use of checklists \cite{ribeiro-etal-2020-beyond}) when the performance of a model is being assessed.

\paragraph{Textbase}
This level covers local relations of propositions in the computational model of reading comprehension.
In the context of NLP, it refers to various types of relations linked between sentences.
These relations not only include the typical relations between sentences (discourse relations) but also the links between entities.
Consequently, this level includes coreference resolution, causality, temporal relations, spatial relations, text structuring relations, logical reasoning, knowledge reasoning, commonsense reasoning, and mathematical reasoning.
We also include multi-hop reasoning \cite{welbl2018constructing} at this level because it does not necessarily require a coherent global representation over a given context.
For studying the generalizability of MRC, \citet{fisch2019mrqa} propose a shared task featuring training and testing on multiple domains.
\citet{talmor2019multiqa} and \citet{khashabi-etal-2020-unifiedqa} also find that training on multiple datasets leads to robust generalization.
However, unless we make sure that datasets require various skills 
with sufficient coverage, it might remain unclear whether we evaluate a model's transferability of the reading comprehension ability.

\paragraph{Situation Model}
This level targets the global structure of propositions in human reading comprehension.
It includes a coherent and situational representation of a given context and its grounding to the non-textual information.
A coherent representation has well-organized sentence-to-sentence transitions \cite{barzilay2008modeling}, which are vital for using procedural and script knowledge \cite{schank1977scripts}.
This level also includes characters' goals and plans, meta perspective including author's intent and attitude, thematic understanding, and grounding to other media.
Most existing MRC datasets seem to struggle to target the situation model.
We discuss further in Section~\ref{sec:what-direction}.

\begin{figure}[!t]
  \centering \small
  \fbox{
    \parbox{0.95\linewidth}{
      \begin{minipage}{0.99\linewidth} \vspace{0.2em}
        Passage: \emph{The princess climbed out the window of the high tower and climbed down the south wall when her mother was sleeping. She wandered out a good way. Finally, she went into the forest where there are no electric poles.} \\

        Q1: \emph{Who climbed out of the castle?} A: \emph{Princess}

        Q2: \emph{Where did the princess wander after escaping?} \\ A: \emph{Forest}

        Q3: \emph{What would happen if her mother was not sleeping?} A: \emph{the princess would be caught soon (multiple choice)}
      \end{minipage}
    }
  }
  \caption{Example questions of the different representation levels. The passage is taken from MCTest.}
  \label{fig:example}
\end{figure}

\paragraph{Example}
\toread\ The representation levels in the example shown in Figure \ref{fig:example} are described as follows.
Q1 is at the surface-structure level where a reader only needs to understand the subject of the first event.
We expect that Q2 requires understanding of relations among described entities and events at the textbase level; the reader may need to understand who \emph{she} means using coreference resolution.
\emph{Escaping} in Q2 also requires the reader's commonsense to associate it with the first event.
However, the reader might be able to answer this question only by looking for a place (specified by \emph{where}) described in the passage, thereby necessitating the validity of the question to correctly evaluate the understanding of the described events.
Q3 is an example that requires imagining a different situation at the situation-model level, which could be further associated with a grounding question such as \emph{which figure best depicts the given passage?}

In summary, we indicate that the following features might be missing in existing datasets:

\begin{itemize}[leftmargin=1.5em] 
\item Considering the capability to acquire basic understanding of the linguistic-level information.
\item Ensuring that the questions comprehensively specify and evaluate textbase-level skills.
\item Evaluating the capability of the situation model in which propositions are coherently organized and are grounded to non-textual information. 
\end{itemize}

\paragraph{Should MRC models mimic human text comprehension?}
\toread\~ In this paper, we do not argue that MRC models should mimic human text comprehension.
However, when we design an NLU task and create datasets for testing human-like linguistic generalization, we can refer to the aforementioned features to frame the intended behavior to evaluate in the task.
As \citet{linzen-2020-accelerate} discusses, the task design is orthogonal to how the intended behavior is realized at the implementation level \cite{marr1982vision}.

\section{MRC on Psychometrics}
\label{sec:how-mrc}

In this section, we provide a theoretical foundation for the evaluation of MRC models.
When MRC measures the capability of reading comprehension, validation of the measurement is crucial to obtain a reliable and useful explanation.
Therefore, we focus on psychometrics---a field of study concerned with the assessment of the quality of psychological measurement \cite{furr2018psychometrics}.
We expect that the insights obtained from psychometrics can facilitate a better task design.
In Section~\ref{sec:psycho-validity}, we first review the concept of validity in psychometrics.
Subsequently, in Section~\ref{sec:mrc-validity}, we examine the aspects that correspond to construct validity in MRC and then indicate the prerequisites for verifying the intended explanation of MRC in its task design.

\subsection{Construct Validity in Psychometrics}
\label{sec:psycho-validity}

\begin{table*}[t]
  \def\arraystretch{1.4} \small
  \begin{tabular}{p{7.5em}p{20em}p{19em}} \toprule
    Validity aspects & Definition in psychometrics & Correspondence in MRC \\ \midrule
    1. Content & Evidence of content relevance, representativeness, and technical quality. & Questions require reading comprehension skills with sufficient coverage and representativeness over the representation levels. \\
    2. Substantive & Theoretical rationales for the observed consistencies in the test responses including task performance of models. 
    & Questions correctly evaluate the intended intermediate process of reading comprehension and provide rationales to the interpreters. \\ 
    3. Structural & Fidelity of the scoring structure to the structure of the construct domain at issue. & Correspondence between the task structure and the score structure. \\ 
    4. Generalizability & Extent to which score properties and interpretations can be generalized to and across population groups, settings, and tasks. & Reliability of test scores in correct answers and model predictions, and applicability to other datasets and models. \\
    5. External & Extent to which the assessment scores' relationship with other measures and non-assessment behaviors reflect the expected relations. & Comparison of the performance of MRC with that of other NLU tasks and measurements. \\  
    6. Consequential & Value implications of score interpretation as a basis for 
    the consequences of test use, especially regarding the sources of invalidity related to issues of bias, fairness, and distributive justice.
    & Considering the model vulnerabilities to adversarial attacks and social biases of models and datasets to ensure the fairness of model outputs. \\
    \bottomrule
  \end{tabular}
  \caption{
    Aspects of the construct validity in psychometrics and corresponding features in MRC.
  }
  \label{tab:validity}
\end{table*}

According to psychometrics, \emph{construct validity} is necessary to validate the interpretation of outcomes of psychological experiments.\footnote{In psychology, a construct is an abstract concept, which facilitates the understanding of human behavior such as vocabulary, skills, and comprehension.}
\citet{messick1995validity} report that construct validity consists of the six aspects shown in Table~\ref{tab:validity}.

In the design of educational and psychological measurement, these aspects collectively provide verification questions that need to be answered for justifying the interpretation and use of test scores.
In this sense, the construct validation can be viewed as an empirical evaluation of the meaning and consequence of measurement.
Given that MRC is intended to capture the reading comprehension ability, the task designers need to be aware of these validity aspects.
Otherwise, users of the task cannot justify the score interpretation, i.e., it cannot be confirmed that successful systems actually perform intended reading comprehension.

\subsection{Construct Validity in MRC}
\label{sec:mrc-validity}

Table~\ref{tab:validity} also raises MRC features corresponding to the six aspects of construct validity.
In what follows, we elaborate on these correspondings and discuss the missing aspects that are needed to achieve the construct validity of the current MRC. 

\paragraph{Content Aspect}
As discussed in Section~\ref{sec:what-mrc}, sufficiently covering the skills across all the representation levels is an important requirement of MRC.
It may be desirable that an MRC model is simultaneously evaluated on various skill-oriented examples.

\paragraph{Substantive Aspect}
This aspect appraises the evidence for the consistency of model behavior.
We consider that this is the most important aspect for explaining reading comprehension, a process that subsumes various implicit and complex steps.
To obtain a consistent response from an MRC system, it is necessary to ensure that the questions correctly assess the internal steps in the process of reading comprehension.
However, as stated in Section~\ref{sec:issues}, most existing datasets fail to verify that a question is solved by using an intended skill, which implies that it cannot be proved that a successful system can actually perform intended comprehension.

\paragraph{Structural Aspect}
Another issue in the current MRC is that they only provide simple accuracy as a metric.
Given that the substantive aspect necessitates the evaluation of the internal process of reading comprehension, the structure of metrics needs to reflect it.
However, a few studies have attempted to provide a dataset with multiple metrics.
For example, \citet{yang2018hotpotqa} not only ask for the answers to questions but also provide sentence-level supporting facts.
This metric can also evaluate the process of multi-hop reasoning whenever the supporting sentences need to be understood for answering a question.
Therefore, we need to consider both substantive and structural aspects.

\paragraph{Generalizability Aspect}
The generalizability of MRC can be understood from the reliability of metrics and the reproducibility of findings.
For the reliability of metrics, we need to take care of the reliability of gold answers and model predictions.
Regarding the accuracy of answers, the performance of the model becomes unreliable when the answers are unintentionally ambiguous or impractical.
Because the gold answers in most datasets are only decided by the majority vote of crowd workers, the ambiguity of the answers is not considered.
It may be useful if such ambiguity can be reflected in the evaluation (e.g., using the item response theory \cite{lalor2016building}).
As for model predictions, an issue may be the reproducibility of results \cite{bouthillier2019unreproducible}, which implies that the reimplementation of a system generates statistically similar predictions.
For the reproducibility of models, \citet{dror2018hitchhikers} emphasize statistical testing methods to evaluate models.
For the reproducibility of findings, \citet{bouthillier2019unreproducible} stress it as the transferability of findings 
in a dataset/task to another dataset/task.
In open-domain question answering, \citet{lewis2021question} point out that successful models might only memorize dataset-specific knowledge.
To facilitate this transferability, we need to have 
units of explanation that can be used in different datasets \cite{doshi2018considerations}.

\paragraph{External Aspect}
This aspect refers to the relationship between a model's scores on different tasks.
\citet{yogatama2019learning} point out that current models struggle to transfer their ability from a task originally trained on (e.g., MRC) to different unseen tasks (e.g., SRL).
To develop a general NLU model, one would expect that a successful MRC model should show sufficient performance on other NLU tasks as well.
To this end, \citet{wang2019superglue} propose an evaluation framework with ten different NLU tasks in the same format. 

\paragraph{Consequential Aspect}
This aspect refers to the actual and potential consequences of test use.
In MRC, this refers to the use of a successful model in practical situations other than tasks, where we need to ensure the robustness of a model to adversarial attacks and the accountability for unintended model behaviors.
\citet{wallace2019universal} highlight this aspect by showing that existing NLP models are vulnerable to adversarial examples, thereby generating egregious outputs.

\paragraph{Summary: Design of Rubric}\toread\
Given the validity aspects, our suggestion is to design a \emph{rubric} (scoring guide used in education) of what reading comprehension we expect is evaluated in a dataset; this helps to inspect detailed strengths and weaknesses of models that cannot be obtained only by simple accuracy.
The rubric should not only cover various linguistic phenomena (the \emph{content} aspect) but also involve different levels of intermediate evaluation in the reading comprehension process (the \emph{substantive} and \emph{structural} aspects) as well as stress testing of adversarial attacks (the \emph{consequential} aspect).
The rubric is in a similar motivation with dataset statements \cite{bender-friedman-2018-data,gebru2018datasheets}; however, taking the validity aspects into account would improve its substance.

\section{Future Directions}
\label{sec:direction}

This section discusses future potential directions toward answering the \emph{what} and \emph{how} questions in Sections~\ref{sec:what-mrc} and \ref{sec:how-mrc}.
In particular, we infer that the \emph{situation model} and \emph{substantive validity} are critical for benchmarking human-level MRC.

\subsection{\emph{What} Question: Situation Model}
\label{sec:what-direction}

As mentioned in Section~\ref{sec:what-mrc}, existing datasets fail to fully assess the ability of creating the situation model.
As a future direction, we suggest that the task should deal with two features of the situation model: context dependency and grounding. 

\subsubsection{Context-dependent Situations}
A vital feature of the situation model is that it is conditioned on a given text, i.e., a representation is constructed distinctively depending on the given context.
We elaborate it by discussing the two key features: defeasibility and novelty.

\paragraph{Defeasibility}
The defeasibility of a constructed representation implies that a reader can modify and revise it according to the newly acquired information \cite{davis2015commonsense,schubert2015whatkinds}.
The defeasibility of NLU has been tackled in the task of if-then reasoning \cite{sap2019atomic}, abductive reasoning \cite{bhagavatula2020abductive}, counterfactual reasoning \cite{qin2019counterfactual}, or contrast sets \cite{gardner-etal-2020-evaluating}.
\toread\ A possible approach in MRC is that we ask questions against a set of modified passages that describe slightly different situations, where the same question can lead to different conclusions.

\paragraph{Novelty}
An example showing the importance of contextual novelty is \emph{Could a crocodile run a steeplechase?} by \citet{levesque2014behaviour}.
This question poses a novel situation where the solver needs to combine multiple commonsense knowledge to derive the correct answer.
If non-fiction documents, such as newspaper and Wikipedia articles, are only used, some questions require only the reasoning of facts already known in web-based corpus.
Fictional narratives may be a better source for creating questions on novel situations.


\subsubsection{Grounding to Other Media}
In MRC, grounding texts to non-textual information is not fully explored yet.
\citet{kembhavi2017smarter} propose a dataset based on science textbooks, which contain questions with passages, diagrams, and images.
\citet{ebrahimi2018figureqa} propose a figure-based question answering dataset that requires the understanding of figures including line plots and bar charts.
Although another approach could be vision-based question answering tasks \cite{antol2015VQA,zellers2019recognition}, we cannot directly use them for evaluating NLU because they focus on understanding of images rather than texts.
Similarly to the textbook questions \cite{kembhavi2017smarter}, a possible approach would be to create questions for understanding of texts through showing figures.
\toread\ We might also need to account for the \emph{scope} of grounding \cite{bisk2020experience}, i.e., ultimately understanding human language in a social context beyond simply associating texts with perceptual information.

\subsection{\emph{How} Question: Substantive Validity}
\label{sec:how-direction}
Substantive validity requires us to ensure that the questions correctly assess the internal steps of reading comprehension. 
We discuss two approaches for this challenge: creating shortcut-proof questions and ensuring the explanation by design.

\subsubsection{Shortcut-proof Questions}

\citet{gururangan2018annotation} reveal that NLU datasets can contain unintended dataset biases embedded by annotators.
If machine learning models exploit such biases for answering questions, we cannot evaluate the precise NLU of models.
Following \citet{geirhos2020shortcut}, we define \emph{shortcut-proof} questions as ones that prevent models from exploiting dataset biases and learning decision rules (\emph{shortcuts}) that perform well only on i.i.d. test examples with regard to its training examples.
\citet{gardner-etal-2019-making} also point out the importance of mitigating shortcuts in MRC.
In this section, we view two different approaches for this challenge. 

\paragraph{Removing Unintended Biases by Filtering}

\citet{zellers2018swag} propose a model-based adversarial filtering method that iteratively trains an ensemble of stylistic classifiers and uses them to filter out the questions.
\citet{sakaguchi2019winogrande} also propose filtering methods based on both machines and humans to alleviate \emph{dataset-specific} and \emph{word-association} biases.
However, a major issue is the inability to discern knowledge from bias in a closed domain.
When the domain is equal to a dataset, patterns that are valid only in the domain are called \emph{dataset-specific} biases (or annotation artifacts in the labeled data).
When the domain covers larger corpora, the patterns (e.g., frequency) are called \emph{word-association} biases.
When the domain includes everyday experience, patterns are called \emph{commonsense}.
However, as mentioned in Section~\ref{sec:what-direction}, commonsense knowledge can be \emph{defeasible}, which implies that the knowledge can be false in unusual situations.
In contrast, when the domain is our real world, indefeasible patterns are called \emph{factual knowledge}.
Therefore, the distinction of bias and knowledge depends on where the pattern is recognized.
This means that a dataset should be created so that it can evaluate reasoning on the intended knowledge.
For example, to test defeasible reasoning, we must filter out questions that are solvable by usual commonsense only.
If we want to investigate the reading comprehension ability without depending on factual knowledge, we can consider counterfactual or fictional situations.

\paragraph{Identifying Requisite Skills by Ablating Input Features}

Another approach is to verify shortcut-proof questions by analyzing the human answerability of questions regarding their key features.
We speculate that if a question is still answerable by humans even after removing the intended features, the question does not require understanding of the ablated features  (e.g., checking the necessity of resolving pronoun coreference after replacing pronouns with dummy nouns).
Even if we cannot accurately identify such necessary features, by identifying partial features of them in a sufficient number of questions, we could expect that the questions evaluate the corresponding intended skill.
In a similar vein, \citet{geirhos2020shortcut} argue that a dataset is useful only if it is a good proxy for the underlying ability one is actually interested in.

\subsubsection{Explanation by Design}
\label{sec:white-box}
Another approach for ensuring the substantive validity is to include explicit explanation in the task formulation.
Although gathering human explanations is costly, the following approaches can facilitate the explicit verification of a model's understanding using a few test examples.

\paragraph{Generating Introspective Explanation}
\citet{inoue2020r4c} classify two types of explanation in text comprehension: \emph{justification explanation} and \emph{introspective explanation}.
The justification explanation only provides a collection of supporting facts for making a certain decision, whereas the introspective explanation provides the derivation of the answer for making the decision, which can cover linguistic phenomena and commonsense knowledge not explicitly mentioned in the text.
They annotate multi-hop reasoning questions with introspective explanation and propose a task that requires the derivation of the correct answer of a given question to improve the explainability.
\citet{rajani2019explain} collect human explanations for commonsense reasoning and improve the system's performance by modeling the generation of the explanation.
\toread\ Although we must take into account the faithfulness of explanation, asking for introspective explanations could be useful in inspecting the internal reasoning process, e.g., by extending the task formulation so that it includes auxiliary questions that consider the intermediate facts in a reasoning process.
For example, before answering Q2 in Figure \ref{fig:example}, a reader should be able to answer \emph{who escaped?} and \emph{where did she escape from?} at the surface-structure level.

\paragraph{Creating Dependency Between Questions}
Another approach for improving the substantive validity is to create dependency between questions by which answering them correctly involves answering some other questions correctly.
For example, \citet{dalvi2018tracking} propose a dataset that requires a procedural understanding of scientific facts.
In their dataset, a set of questions corresponds to the steps of the entire process of a scientific phenomenon.
Therefore, this set can be viewed as a single question that requires a complete understanding of the scientific phenomenon.
In CoQA \cite{reddy2019coqa}, it is noted that questions often have pronouns that refer back to nouns appearing in previous questions.
\toread\ These mutually-dependent questions can probably facilitate the explicit validation of the models' understanding of given texts.

\section{Conclusion}

In this paper, we outlined current issues and future directions for benchmarking machine reading comprehension.
We visited the psychology study to analyze \emph{what} we should ask of reading comprehension and the construct validity in psychometrics to analyze \emph{how} we should correctly evaluate it.
We deduced that future datasets should evaluate the capability of the situation model for understanding context-dependent situations and for grounding to non-textual information and ensure the substantive validity by creating shortcut-proof questions and designing an explanatory task formulation.

\section*{Acknowledgments}
The authors would like to thank Xanh Ho for helping create the dataset list and the anonymous reviewers for their insightful comments.
This work was supported by JSPS KAKENHI Grant Number 18H03297, JST ACT-X Grant Number JPMJAX190G, and JST PRESTO Grant Number JPMJPR20C4.

\bibliography{eacl2021}
\bibliographystyle{acl_natbib}


\appendix

\newcommand{\datasetcaption}{In the \emph{answer style} column, \emph{descript} represents description (free-form answering) and \emph{extract} denotes answer extraction by selecting a span in given texts. \emph{Size} indicates the size of the whole dataset including training, development, and test sets. In the \emph{question source} column, \emph{crowd} indicates questions written by crowdworkers and \emph{query} indicates questions collected from search-engine queries.}

\begin{table*}[!b]\footnotesize\centering
  \def\arraystretch{2.0}
  \setlength{\tabcolsep}{10pt}
  \input{dataset_table1}
  \caption{Machine reading comprehension datasets published until 2017. \datasetcaption}
  \label{tbl:dataset1}
\end{table*}

\begin{table*}[th]\footnotesize\centering
  \def\arraystretch{2.0}
  \setlength{\tabcolsep}{10pt}
  \input{dataset_table2}
  \caption{Machine reading comprehension datasets published in 2018. \datasetcaption}
  \label{tbl:dataset2}
\end{table*}

\begin{table*}[th]\footnotesize\centering
  \def\arraystretch{2.0}
  \setlength{\tabcolsep}{10pt}
  \input{dataset_table3}
  \caption{Machine reading comprehension datasets published in 2019. \datasetcaption}
  \label{tbl:dataset3}
\end{table*}

\begin{table*}[th]\footnotesize\centering
  \def\arraystretch{2.0}
  \setlength{\tabcolsep}{10pt}
  \input{dataset_table4}
  \caption{Machine reading comprehension datasets published in 2020. \datasetcaption}
  \label{tbl:dataset4}
\end{table*}

\section{Machine Reading Comprehension Datasets}
\label{app:datasets}

Tables \ref{tbl:dataset1}, \ref{tbl:dataset2}, \ref{tbl:dataset3}, and \ref{tbl:dataset4} list machine reading comprehension and related datasets along with their answer styles, dataset size, type of corpus, sourcing methods, and focuses.

\end{document}

%% file: skill-hierarchy.tex
\begin{tikzpicture}
  \small
\coordinate (A) at (-2.5, 0) {};
\coordinate (B) at ( 2.5, 0) {};
\coordinate (C) at (0, 3.5) {};
\draw[name path=AC] (A) -- (C);
\draw[name path=BC] (B) -- (C);

\path[name path=horiz] (A|-0,2) -- (B|-0,2);
\draw[name intersections={of=AC and horiz,by=P},
  name intersections={of=BC and horiz,by=Q}] (P) -- (Q)
node[midway,above](X3) {\shortstack{Situation \\ model}};

\path[name path=horiz] (A|-0,1) -- (B|-0,1);
\draw[name intersections={of=AC and horiz,by=P},
  name intersections={of=BC and horiz,by=Q}] (P) -- (Q)
node[midway,above=0.8em](X2) {Textbase};

\path[name path=horiz] (A|-0,0) -- (B|-0,0);
\draw[name intersections={of=AC and horiz,by=P},
  name intersections={of=BC and horiz,by=Q}] (P) -- (Q)
node[midway,above=0.8em](X1) {Surface structure};


\newcommand{\boxsize}{33em}

\node[draw,above=2.0em, text width=\boxsize](N3) at (8,2){\small
  \begin{tabular}{l}
    Construct the global structure of propositions. \\
     \textit{Skills: creating a coherent representation and grounding it to other media.} \\
  \end{tabular}
};
\node[draw,above=0.4em, text width=\boxsize](N2) at (8,1){\small
  \begin{tabular}{p{31em}}
    Construct the local relations of propositions. \\
    \textit{Skills: recognizing relations between sentences such as coreference resolution, knowledge reasoning, and understanding discourse relations.} \\
  \end{tabular}
}; 

\node[draw,above=-0.3em, text width=\boxsize](N1) at (8,0){\small
  \begin{tabular}{l}
    Creating propositions from the textual input. \\
    \textit{Skills: syntactic and dependency parsing, POS tagging, SRL, and NER.} \\
  \end{tabular}
};
\draw[->,dotted] (X3) -- (N3.west);
\draw[->,dotted] (X2) -- (N2.west);
\draw[->,dotted] (X1) -- (N1.west);
\end{tikzpicture}

%% file: dataset_table1.tex
\begin{tabular}{ccccccccccccc} \toprule
Dataset & \linestack{Answer \\ style} & Size & Corpus & \linestack{Question \\ source} & \linestack{Focus} \\ \midrule 
 \linestack{QA4MRE\\\cite{sutcliffe2013QA4MRE}} & \linestack{multiple- \\ choice} & 240 & \linestack{technical \\ document} & expert & \linestack{exam-level questions} \\ 
 \linestack{MCTest\\\cite{richardson2013mctest}} & \linestack{multiple- \\ choice} & 2.6K & \linestack{written \\ story} & crowd & \linestack{children-level narrative} \\ 
 \linestack{bAbI\\\cite{weston2015bAbI}} & descript & \linestack{10K * \\ 20} & \linestack{generated \\ text} & automated & \linestack{toy tasks for prerequisite skills} \\ 
 \linestack{CNN/ DailyMail\\\cite{hermann2015teaching}} & extract & 1.4M & \linestack{news \\ article} & automated & \linestack{entity cloze} \\ 
 \linestack{Children's Book Test\\\cite{hill2015goldilocks}} & extract & 688K & narrative & automated & \linestack{large-scale automated} \\ 
 \linestack{SQuAD 1.1\\\cite{rajpurkar2016squad}} & extract & 100K & Wikipedia & crowd & \linestack{large-scale crowdsourced} \\ 
 \linestack{LAMBADA\\\cite{paperno2016lambada}} & descript & 10K & narrative & crowd & \linestack{hard language modeling} \\ 
 \linestack{WikiReading\\\cite{hewlett2016wikireading}} & descript & 18m & Wikipedia & automated & \linestack{Wikidata articles} \\ 
 \linestack{Who did What\\\cite{onishi2016who}} & \linestack{multiple- \\ choice} & 200K & \linestack{news \\ article} & automated & \linestack{cloze of person names} \\ 
 \linestack{MS MARCO\\\cite{nguyen2016msmarco}} & descript & 100K & \linestack{web \\ snippet} & query & \linestack{description on web snippets} \\ 
 \linestack{NewsQA\\\cite{trischler2017newsqa}} & extract & 120K & \linestack{news \\ article} & crowd & \linestack{blindly created questions} \\ 
 \linestack{SearchQA\\\cite{dunn2017searchqa}} & extract & 140K & \linestack{web \\ snippet} & trivia & \linestack{49.6 snippets on average} \\ 
 \linestack{RACE\\\cite{lai2017race}} & \linestack{multiple- \\ choice} & 100K & \linestack{language \\ exam} & expert & \linestack{middle and high school \\ English exam in China} \\ 
 \linestack{Story Cloze Test\\\cite{mostafazadeh2017lsdsem}} & \linestack{multiple- \\ choice} & 3.7K & \linestack{written \\ story} & crowd & \linestack{98,159 stories for training} \\ 
 \linestack{TriviaQA\\\cite{joshi2017triviaqa}} & extract & 650K & \linestack{web \\ snippet} & trivia & \linestack{trivia questions} \\ 
 \linestack{Quasar\\\cite{dhingra2017quasar}} & extract & 80K & \linestack{web \\ snippet} & query & \linestack{search queries} \\ 
 \linestack{TextbookQA\\\cite{kembhavi2017smarter}} & \linestack{multiple- \\ choice} & 26K & textbook & expert & \linestack{figures included} \\ 
 \linestack{AddSent SQuAD\\\cite{jia2017adversarial}} & extract & 3.6K & Wikipedia & crowd & \linestack{distracting sentences injected} \\
\bottomrule \end{tabular}

%% file: dataset_table2.tex
\begin{tabular}{ccccccccccccc} \toprule
Dataset & \linestack{Answer \\ style} & Size & Corpus & \linestack{Question \\ source} & \linestack{Focus} \\ \midrule 
 \linestack{ARCT\\\cite{habernal2018argument}} & \linestack{multiple- \\ choice} & 2.0K & \linestack{debate \\ article} & \linestack{crowd \\ expert} & \linestack{reasoning on argument} \\ 
 \linestack{QAngaroo\\\cite{welbl2018constructing}} & \linestack{multiple- \\ choice} & 50K & \linestack{Wikipedia, \\ MEDLINE} & automated & \linestack{multi-hop reasoning} \\ 
 \linestack{CLOTH\\\cite{xie2018large}} & \linestack{multiple- \\ choice} & 99K & various & expert & \linestack{cloze in exam texts} \\ 
 \linestack{NarrativeQA\\\cite{kocisky2018narrativeqa}} & descript & 45K & \linestack{movie \\ script} & crowd & \linestack{summary and full \\ story tasks} \\ 
 \linestack{MCScript\\\cite{ostermann2018mcscript}} & \linestack{multiple- \\ choice} & 30K & \linestack{written \\ story} & crowd & \linestack{commonsense reasoning, \\ script knowledge} \\ 
 \linestack{CliCR\\\cite{suster2018clicr}} & extract & 100K & \linestack{clinical case \\ text} & automated & \linestack{cloze style queries} \\ 
 \linestack{ARC\\\cite{clark2018think}} & \linestack{multiple- \\ choice} & 8K & \linestack{science \\ exam} & expert & \linestack{retrieved documents \\ from textbooks} \\ 
 \linestack{DuoRC\\\cite{saha2018duorc}} & extract & 186K & \linestack{movie \\ script} & crowd & \linestack{commonsense reasoning, \\ multi-sentence reasoning} \\ 
 \linestack{ProPara\\\cite{dalvi2018tracking}} & extract & 2K & \linestack{science \\ exam} & automated & \linestack{procedural understanding} \\ 
 \linestack{DuReader\\\cite{he2018dureader}} & descript & 200K & \linestack{web \\ snippet} & \linestack{query \\ crowd} & \linestack{Chinese, \\ Baidu Search/Knows} \\ 
 \linestack{MultiRC\\\cite{khashabi2018looking}} & \linestack{multiple- \\ choice} & 6K & \linestack{various \\ documents} & crowd & \linestack{multi-sentence reasoning} \\ 
 \linestack{Multi-party Dialog\\\cite{ma2018challenging}} & extract & 13K & \linestack{TV show \\ transcript} & automated & \linestack{1.7k crowd dialogues, \\ cloze query} \\ 
 \linestack{SQuAD 2.0\\\cite{rajpurkar2018know}} & \linestack{extract \\ no answer} & 100K & Wikipedia & crowd & \linestack{unanswerable questions} \\ 
 \linestack{ShARC\\\cite{saeidi2018interpretation}} & \linestack{yes/no/ \\ irrelevant} & 32K & \linestack{web \\ snippet} & crowd & \linestack{reasoning on rules from \\ government documents} \\ 
 \linestack{QuAC\\\cite{choi2018quac}} & \linestack{extract \\ yes/no} & 100K & Wikipedia & crowd & \linestack{dialogue-based, \\ 14k dialogues} \\ 
 \linestack{Textworlds QA\\\cite{labutov2018multi}} & extract & 1.2M & \linestack{generated \\ text} & automated & \linestack{simulated worlds, \\ logical reasoning} \\ 
 \linestack{SWAG\\\cite{zellers2018swag}} & \linestack{multiple- \\ choice} & 113K & \linestack{video \\ captions} & \linestack{language- \\ model} & \linestack{commonsense reasoning} \\ 
 \linestack{emrQA\\\cite{pampari2018emrqa}} & extract & 400K & \linestack{clinical \\ documents} & automated & \linestack{using annotated logical \\ forms on i2b2 dataset} \\ 
 \linestack{HotpotQA\\\cite{yang2018hotpotqa}} & \linestack{extract \\ yes/no} & 113K & Wikipedia & crowd & \linestack{multi-hop reasoning} \\ 
 \linestack{OpenbookQA\\\cite{mihaylov2018suit}} & \linestack{multiple- \\ choice} & 6.0K & textbook & crowd & \linestack{commonsense reasoning} \\ 
 \linestack{RecipeQA\\\cite{yagcioglu2018recipeqa}} & \linestack{multiple- \\ choice} & 36K & \linestack{recipe \\ script} & automated & \linestack{multimodal questions} \\ 
 \linestack{ReCoRD\\\cite{zhang2018record}} & extract & 120K & \linestack{news \\ article} & crowd & \linestack{commonsense reasoning, \\ cloze query} \\
\bottomrule \end{tabular}

%% file: dataset_table3.tex
\begin{tabular}{ccccccccccccc} \toprule
Dataset & \linestack{Answer \\ style} & Size & Corpus & \linestack{Question \\ source} & \linestack{Focus} \\ \midrule 
 \linestack{CoQA\\\cite{reddy2019coqa}} & \linestack{extract \\ yes/no} & 127K & Wikipedia & crowd & \linestack{dialogue-based, \\ 8k dialogues} \\ 
 \linestack{Commonsense QA\\\cite{talmor2019commonsenseqa}} & \linestack{multiple- \\ choice} & 12K & ConceptNet & crowd & \linestack{commonsense reasoning} \\ 
 \linestack{Natural Questions\\\cite{kwiatkowski2019natural}} & \linestack{extract \\ yes/no} & 323K & Wikipedia & \linestack{query \\ crowd} & \linestack{short or long answer styles} \\ 
 \linestack{DREAM\\\cite{sun2019dream}} & \linestack{multiple- \\ choice} & 10K & \linestack{language \\ exam} & expert & \linestack{dialogue-based, \\ 6.4k multi-party dialogues} \\ 
 \linestack{DROP\\\cite{dua2019drop}} & descript & 96K & Wikipedia & crowd & \linestack{discrete reasoning} \\ 
 \linestack{SocialIQA\\\cite{sap-etal-2019-social}} & \linestack{multiple- \\ choice} & 38K & crowd & crowd & \linestack{commonsense reasoning \\ about social situation} \\ 
 \linestack{BoolQ\\\cite{clark2019boolq}} & yes/no & 16K & Wikipedia & \linestack{query \\ crowd} & \linestack{boolean questions, \\ subset of Natural Questions} \\ 
 \linestack{MSCript 2.0\\\cite{ostermann2019mcscript2}} & \linestack{multiple- \\ choice} & 20K & narrative & crowd & \linestack{commonsense reasoning, \\ script knowledge} \\ 
 \linestack{HellaSWAG\\\cite{zellers2019hellaswag}} & \linestack{multiple- \\ choice} & 70K & \linestack{web \\ snippet} & \linestack{language- \\ model} & \linestack{commonsense reasoning, \\ WikiHow and ActivityNet} \\ 
 \linestack{CODAH\\\cite{chen-etal-2019-codah}} & \linestack{multiple- \\ choice} & 2.8K & \linestack{written \\ prompt} & expert & \linestack{adversarial collection} \\ 
 \linestack{Quoref\\\cite{dasigi2019quoref}} & extract & 24K & Wikipedia & crowd & \linestack{coreference resolution} \\ 
 \linestack{CosmosQA\\\cite{huang2019cosmos}} & \linestack{multiple- \\ choice} & 36K & narrative & crowd & \linestack{commonsense reasoning} \\ 
 \linestack{PubMedQA\\\cite{jin2019pubmedqa}} & yes/no & 273.5K & PubMed & \linestack{expert \\ automated} & \linestack{biomedical domain, \\ 1k expert questions} \\ 
 \linestack{ROPES\\\cite{lin-etal-2019-reasoning}} & extract & 14K & \linestack{textbook \\ Wikipedia} & crowd & \linestack{paragraph effects \\ in situations} \\
\bottomrule \end{tabular}

%% file: dataset_table4.tex
\begin{tabular}{ccccccccccccc} \toprule
Dataset & \linestack{Answer \\ style} & Size & Corpus & \linestack{Question \\ source} & \linestack{Focus} \\ \midrule 
 \linestack{QuAIL\\\cite{rogers2020getting}} & \linestack{multiple- \\ choice} & 15K & various & crowd & \linestack{prerequisite real tasks} \\ 
 \linestack{QASC\\\cite{khot-etal-2020-qasc}} & \linestack{multiple- \\ choice} & 10K & textbook & crowd & \linestack{knowledge composition} \\ 
 \linestack{AdversarialQA\\\cite{bartolo2020beat}} & extract & 36K & Wikipedia & crowd & \linestack{adversarial collection} \\ 
 \linestack{ReClor\\\cite{yu2020reclor}} & \linestack{multiple- \\ choice} & 6.1K & exam & expert & \linestack{logical reasoning} \\ 
 \linestack{R$^4$C\\\cite{inoue2020r4c}} & \linestack{extract \\ descript} & 5K & Wikipedia & crowd & \linestack{multi-hop reasoning} \\ 
 \linestack{TechQA\\\cite{castelli-etal-2020-techqa}} & descript & 1.4K & \linestack{tech \\ documents} & crowd & \linestack{tech forum questions} \\ 
 \linestack{LogiQA\\\cite{liu2020logiqa}} & \linestack{multiple- \\ choice} & 8.7K & exam & expert & \linestack{logical reasoning} \\ 
 \linestack{ProtoQA\\\cite{boratko-etal-2020-protoqa}} & descript & 9.8K & \linestack{web \\ snippet} & crowd & \linestack{commonsense reasoning \\ over prototypical sittuations} \\ 
 \linestack{IIRC\\\cite{ferguson-etal-2020-iirc}} & descript & 13K & Wikipedia & crowd & \linestack{incomplete information} \\ 
 \linestack{HybridQA\\\cite{chen-etal-2020-hybridqa}} & extract & 70K & Wikipedia & crowd & \linestack{understanding tabular data} \\ 
 \linestack{TORQUE\\\cite{ning-etal-2020-torque}} & extract & 21K & TempEval-3 & crowd & \linestack{temporal ordering} \\ 
 \linestack{2WikiMultiHopQA\\\cite{ho-etal-2020-constructing}} & \linestack{extract yes/no \\ descript} & 200K & Wikipedia & automated & \linestack{multi-hop reasoning} \\
\bottomrule \end{tabular}